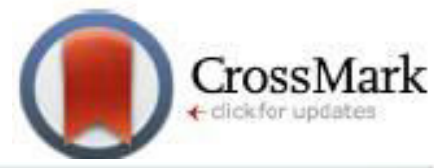

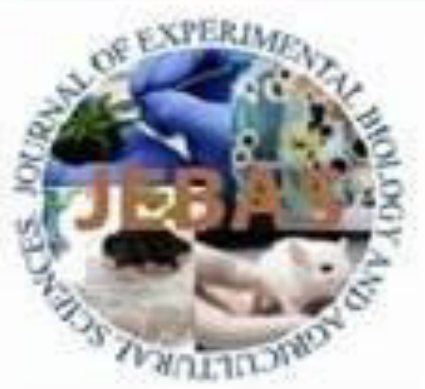

# Integrated Deep Learning Framework Designed on Hybrid Optimization Strategies for Automated Health Detection and Analysis in Silkworms

Komala K V[1]*, Lata B T[2] and Venugopal K R[3]

[1, 2]Computer Science Engineering, University of Visvesvaraya College of Engineering, Bengaluru, India
[3]Computer Science Engineering, Bangalore University, Bengaluru, India

## ABSTRACT

This study proposes a Hybrid Residual-Attention Network (HRAN) for accurately classifying silkworm images into six different classes, including healthy and diseased states. HRAN uses residual blocks for deep feature extraction and attention to focus on disease-related features. A novel Integrated Adaptive Momentum Optimizer (IAMO) was introduced to enhance convergence and improve training efficiency. The dataset of silkworm images underwent preprocessing techniques such as normalization, resizing, and noise reduction, along with augmentation strategies to improve data quality and diversity. The proposed HRAN, optimized using IAMO, achieved an accuracy of 98.67%, outperforming state-of-the-art models such as YOLOv8 and CA-YOLOv5, as demonstrated in benchmarking experiments. The integration of spatial and channel-wise attention mechanisms, coupled with IAMO, significantly enhanced the model's ability to recognize subtle differences between classes. Results indicate that HRAN can be used to detect disease at an early stage in sericulture, and future work will enhance scalability and efficiency in different environments.

* Corresponding author
E-mail: koomlakv2020@gmail.com (Komala K V)

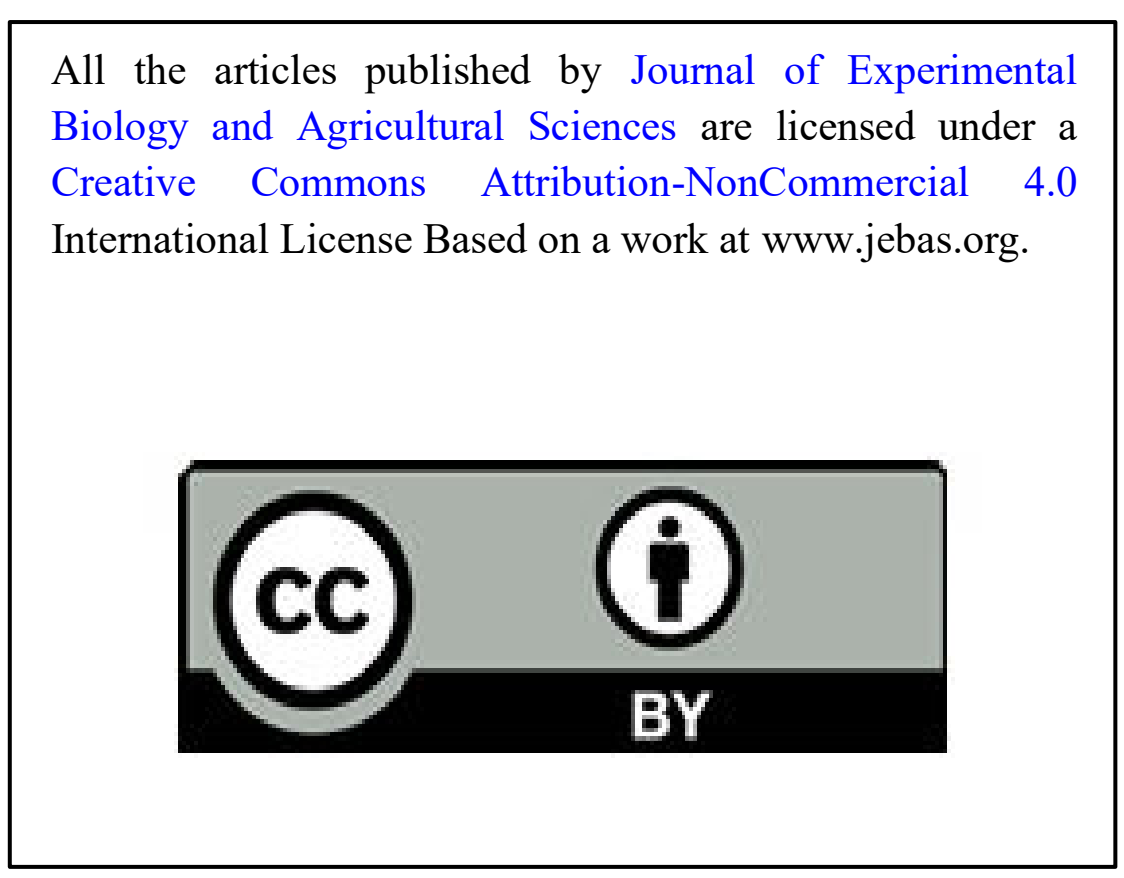

## 1 Introduction

Sericulture is essential for silk production and requires healthy silkworms in large numbers. The health of silkworms directly affects the quantity and quality of silk; hence, disease prevention is a major priority in this industry. Various diseases caused by protozoa, bacteria, fungi, and viruses can affect silkworms. If these diseases go unnoticed and are left uncontrolled, there can be major losses. Conventional approaches to disease diagnosis may rely on skilled manual inspection, which is time-consuming and increases the risk of errors. Large-scale silkworm farms thus offer particularly significant challenges for this strategy (Shi et al., 2023; Zhang et al., 2024). Deep learning presents promising possibilities for automating disease identification and ensuring faster, more accurate findings as technology advances (Zhao et al., 2015). Recent advances in AI-assisted image analysis demonstrate the potential of automated systems for accurate disease detection and classification (Saleh et al., 2025). In the field of agriculture, deep-learning methods have also shown great promise for image-based classification, especially for the extraction of discriminative visual features from complex biological images (B & H P, 2026). This helps to extend the use of advanced deep-learning architectures to automated classification tasks, but the performance of disease-specific classification is dependent on the nature and diversity of the target dataset. In recent years, several researchers have explored deep learning models for plant disease detection, pest identification, and livestock disease prediction, highlighting the effectiveness of automated systems in agriculture and sericulture (Xia et al., 2019). However, limited research has focused on silkworm health monitoring, leaving a critical gap in automated disease detection in this industry.

Recent studies have used deep learning and machine learning for silkworm disease detection. Some methods add attention layers to focus on disease features. IoT-based systems are also used for field monitoring. A novel approach utilizing an enhanced YOLOv8 model incorporated the NAM attention mechanism and ODConv, significantly improved feature extraction and detection speeds, achieving 22.6 milliseconds per image. Notwithstanding these developments, computational complexity poses challenges for deployment in resource-constrained environments; moreover, the model's generalizability is limited by dataset quality (Zhang et al., 2024). Another work presented a YOLOv5s-based architecture including CBAM attention, which improved accuracy by 1.5%.

Still, two recognized limitations were increasing computational load and insufficient situational resilience (Shi et al., 2023). In mixed environments, an improved detection model surpassed conventional techniques in both speed and accuracy through a coordinate attention mechanism. Still, individual datasets are limited, and increasing complexity affects edge machine functionality, resulting in overfitting (Xia et al., 2019). Combining deep learning with machine learning techniques has shown good sex and species classification of silkworm pupae, therefore enhancing classification accuracy and interpretation. Still, factors such as adaptability and dataset-specific constraints (Patten et al., 2011) hampered its general significance. By providing early intervention tools and a simple interface for farmers, a machine learning-based approach effectively detected Grasserie disease. Notwithstanding these advantages, it was noted that environmental variety and response to new disease pressures were limited (Zhao et al., 2015). Using convolutional neural networks for disease recognition, another approach obtained strong results with data augmentation techniques. But responsiveness to image quality and computing needs highlighted a few barriers for general acceptance (Singla et al., 2023). High-accuracy image analysis techniques offered an informal approach to disease detection, enabling earlier therapies.

Still, two accepted barriers include limited scalability and reliance on stable image quality (Nahiduzzaman et al., 2023). Similarly, a VGG-19-based model showed good performance in silkworm classification but suffered from model complexity and sensitivity to illumination conditions (Cappellozza et al., 2022). With limited data, transfer learning systems showed potential for disease recognition, yet computational efficiency and generalization across different datasets remained challenges (Urbanek Krajnc et al., 2022). Emphasizing practicality in field scenarios, an IoT-based surveillance system integrated with artificial intelligence permitted real-time disease diagnosis. Its broader adoption was hindered by challenges in integrating outdated farming techniques and capacity constraints (Li et al., 2023). By enabling greater feature extraction through attention techniques, the proposed attention-concatenation dense convolutional neural network enhanced silkworm disease recognition accuracy. Compared with standard convolutional neural networks, the model achieved notable efficiency gains. Nevertheless, on large-scale sericulture platforms, the capacity and processing requirements of this method remain issues (Zhen et al., 2020). Using YOLO-based methods achieved greater accuracy and speed for real-time disease identification in sericulture. The combination of YOLO and domain-driven traits solved key issues in small-scale silkworm anomaly detection.

Among the challenges are the need for additional computing power and the lack of extensive testing across numerous environmental settings (Shi et al., 2023). Automated feature extraction for Grasserie identification of illnesses in silkworms used capsule networks, thereby achieving modern precision. This approach shows a firm understanding of spatial hierarchy in sick samples. The main disadvantage of capsule networks is their complexity, which requires significant processing capacity and optimization for deployment.

Transfer learning approaches were employed to diagnose Bombyx mori diseases, leveraging pre-trained models. The study highlighted the prospects of transfer learning in sericulture, reducing training time and resource consumption. However, the reliance on transfer learning could limit adaptability to unique regional disease variants (Mungase et al., 2025). Sericulture disease classification with limited data was enhanced by incorporating transfer learning, achieving robust performance in constrained environments. The framework addressed small-dataset issues effectively but faced limitations in scalability for larger datasets and high-dimensional feature spaces (Mungase et al., 2025).

A hybrid CNN-RNN model was developed to predict silkworm diseases, integrating feature selection via convolutional layers and recurrent layers to capture temporal dependencies. This approach improved disease classification accuracy by capturing time-series patterns in data. However, the model's training complexity and potential overfitting with small datasets were identified as challenges (He et al., 2023). IoT-based systems and AI-based systems enabled real-time disease diagnosis in silkworm rearing by integrating image recognition for practical field applications. The study emphasized scalability and farmer-friendly interfaces. However, a lack of advanced automation for decision-making and a limited focus on data security were noted (Guo et al., 2022).

Effective monitoring of silkworms in cultivation systems, made feasible by machine vision technology, provides a non-invasive means of anomaly detection. These methods demonstrated excellent reliability in controlled settings but had difficulty adapting to outdoor and varied environments (Wen et al., 2022). Methodologies for understandable artificial intelligence shed light on predicting disease decision-making in silkworm farming. The work enhanced the AI system's candor and trust but ran throughout trade-offs between computational efficiency and accessibility (Liu et al., 2024). Using image recognition, a continuous tracking system for the silkworm lifecycle classified stages and detected anomalies with high precision. The method had difficulty integrating with current sericulture processes and adjusting to uncontrolled lighting conditions, despite its accuracy (Wen et al., 2022). The work developed an innovative monitoring system using image recognition techniques to track the silkworm's lifetime. The technology provided real-time sericulture management data, demonstrating high accuracy in distinguishing among many growth stages and detecting problems. The study also faced some limitations. The model had difficulty with images taken under uncontrolled conditions, such as changing light and different backgrounds. It also requires further testing before it can be integrated into regular sericulture practices (Shilpashree et al., 2023).

Leveraging several base learners, an ensemble machine learning model categorized healthy and ill silkworms with increasing accuracy. The model showed resilience across several testing patterns but required optimization to reduce training time and complexity (Singla et al., 2023). Using IoT devices for seamless monitoring, AI-powered systems found pests and diseases in silkworm farming. This approach provided early warning mechanisms, but challenges in broadly distributed data integration and cost-effectiveness were noted (Liu et al., 2008). A convolutional neural network effectively classified silkworm species with high accuracy, thereby presenting insights into genetic diversity. However, the study lacked extensive validation in varied environmental and genetic contexts (Sumriddetchkajorn et al., 2015). High-throughput analysis for disease diagnosis in silkworm farming used advanced AI models to streamline diagnostic processes. This approach reduced manual labor but faced challenges regarding operating costs and the need for specialized hardware (Fu et al., 2023).

Earlier studies have improved silkworm disease detection, but some gaps remain. Many models work well only on specific datasets. Some need high computing power for real-time use. Others focus on accuracy and give less attention to small disease-related visual changes. This study addresses these gaps using a deep learning framework that identifies subtle differences between healthy and diseased silkworm classes while maintaining a stable training process. The diversity and dynamic nature of microbial communities associated with plant surfaces have been emphasized by Arora *et al.* (2025).

This study develops a deep learning framework for detecting silkworm health and classifying diseases. The model uses a Hybrid Residual-Attention Network (HRAN) to classify silkworm images as healthy or diseased. Residual blocks extract deep image features. Spatial and channel-wise attention layers help the model focus on disease-related regions. This helps detect small visual changes across different silkworm disease classes. The study also uses an Integrated Adaptive Momentum Optimizer (IAMO) to improve training stability and convergence. The main contributions are: (i) a lightweight deep learning model for silkworm disease classification; (ii) the use of attention layers for disease-specific feature learning; and (iii) model evaluation on silkworm disease datasets, showing higher accuracy, lower loss, and practical value for sericulture. The remaining sections present the materials and methods, results, discussion, and conclusion.

## 2 Materials and Methods

### 2.1 Dataset preparation

The dataset used in this study comprises images of silkworms categorized into six classes: Flacherie, Grasserie, Healthy, Muscardine, Pebrine, and Sick. Each class represents a different silkworm health condition. Flacherie shows soft, weak body symptoms associated with bacterial or viral infections. Images of healthy silkworms do not show visible disease symptoms. Muscardine is a fungal infection characterized by visible mold growth. Grasserie shows a viral infection with shiny skin and swelling. Pebrine is a protozoan infection characterized by black patches and uneven development. The Sick category contains diseased silkworms that are not included in these specific disease classes.

The dataset is prepared from publicly available silkworm images. The images are manually grouped based on visible health conditions and disease symptoms. Healthy and Sick show the general health condition of silkworms, while Flacherie, Grasserie, Muscardine, and Pebrine show specific disease conditions. Since the original images are limited, augmentation is used to increase image variation. For evaluation, 70% of the dataset is used for training, 20% for validation, and 10% for testing. This dataset preparation follows a study-specific approach, where publicly available silkworm images are manually grouped, augmented, and used for health-condition and disease-class classification.

Preprocessing steps are performed to improve the dataset's reliability and accuracy. All images are resized to 224 × 224 pixels to meet the input requirement of the Hybrid Residual-Attention Network (HRAN). Pixel values are normalized to the range of [0, 1] to improve convergence during training. Basic noise suppression is used to reduce image defects caused by background noise and variations in image quality. Data augmentation is applied to improve the model’s generalizability and robustness. These steps include random rotation, horizontal and vertical flipping, random zooming, brightness adjustment, and Gaussian noise addition. Together, these preprocessing and augmentation steps prepare the dataset for HRAN model training and evaluation. Figure 1 shows the images obtained after preprocessing.

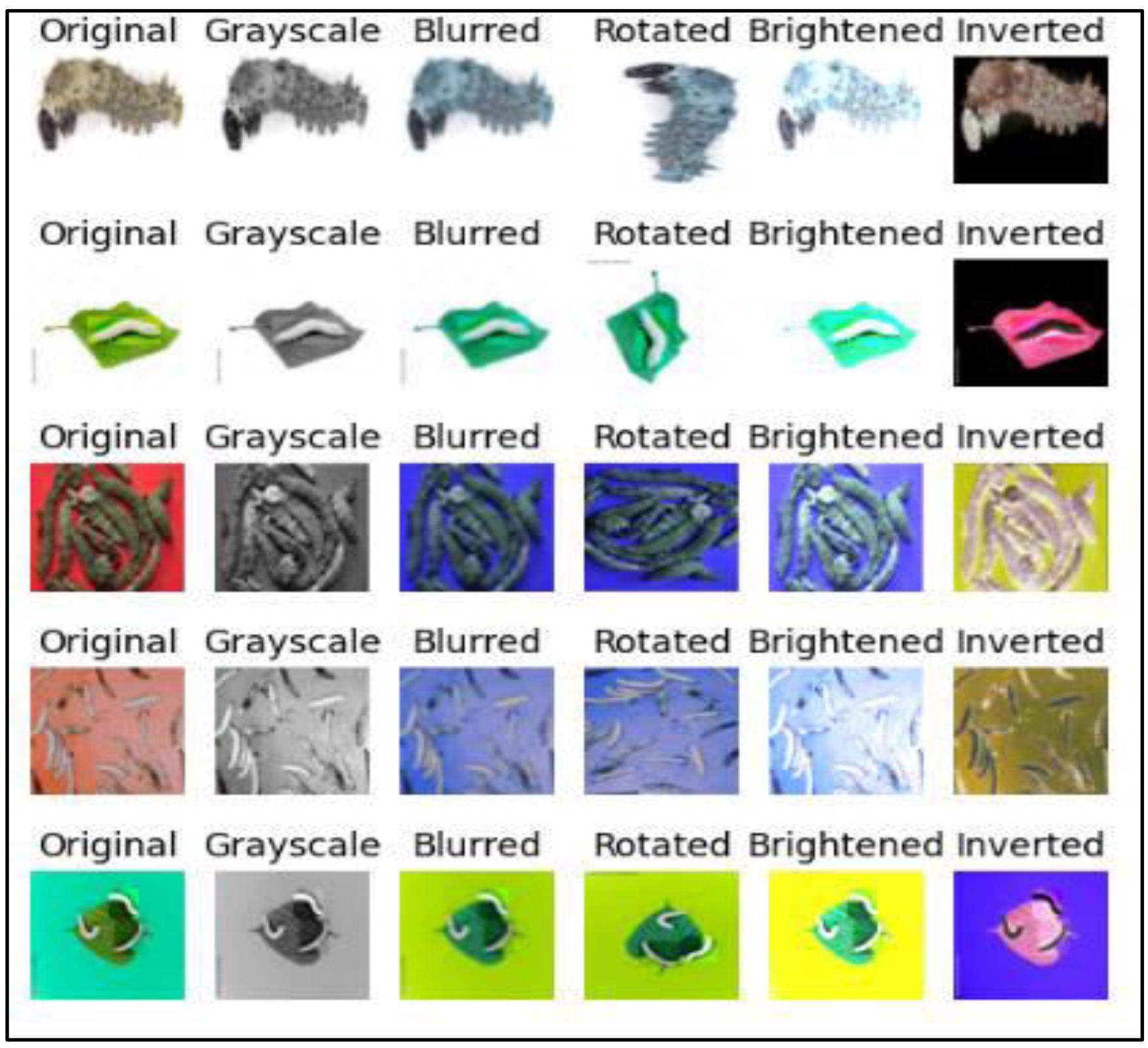

Figure 1 Silkworm images obtained after data preprocessing

## 2.2 Proposed HRAN architecture

The Hybrid Residual-Attention Network (HRAN) is a deep learning model tailored to classify silkworm diseases with high accuracy. The architecture combines residual learning with attention systems to capture disease-specific features while maintaining computational efficiency. HRAN architecture is designed to effectively classify silkworm images into six distinct conditions. The input layer accepts pre-processed silkworm images, resized to a fixed dimension such as 224 × 224 × 3 for RGB images. The backbone of HRAN comprises residual blocks inspired by ResNet, which facilitate deep feature extraction and mitigate vanishing-gradient issues through skip connections. Each residual block contains batch normalization, ReLU activation, convolution layers, and identity mappings. These residual blocks enhance the network's ability to learn complex features without degradation. Figure 2 displays the architecture of HRAN.

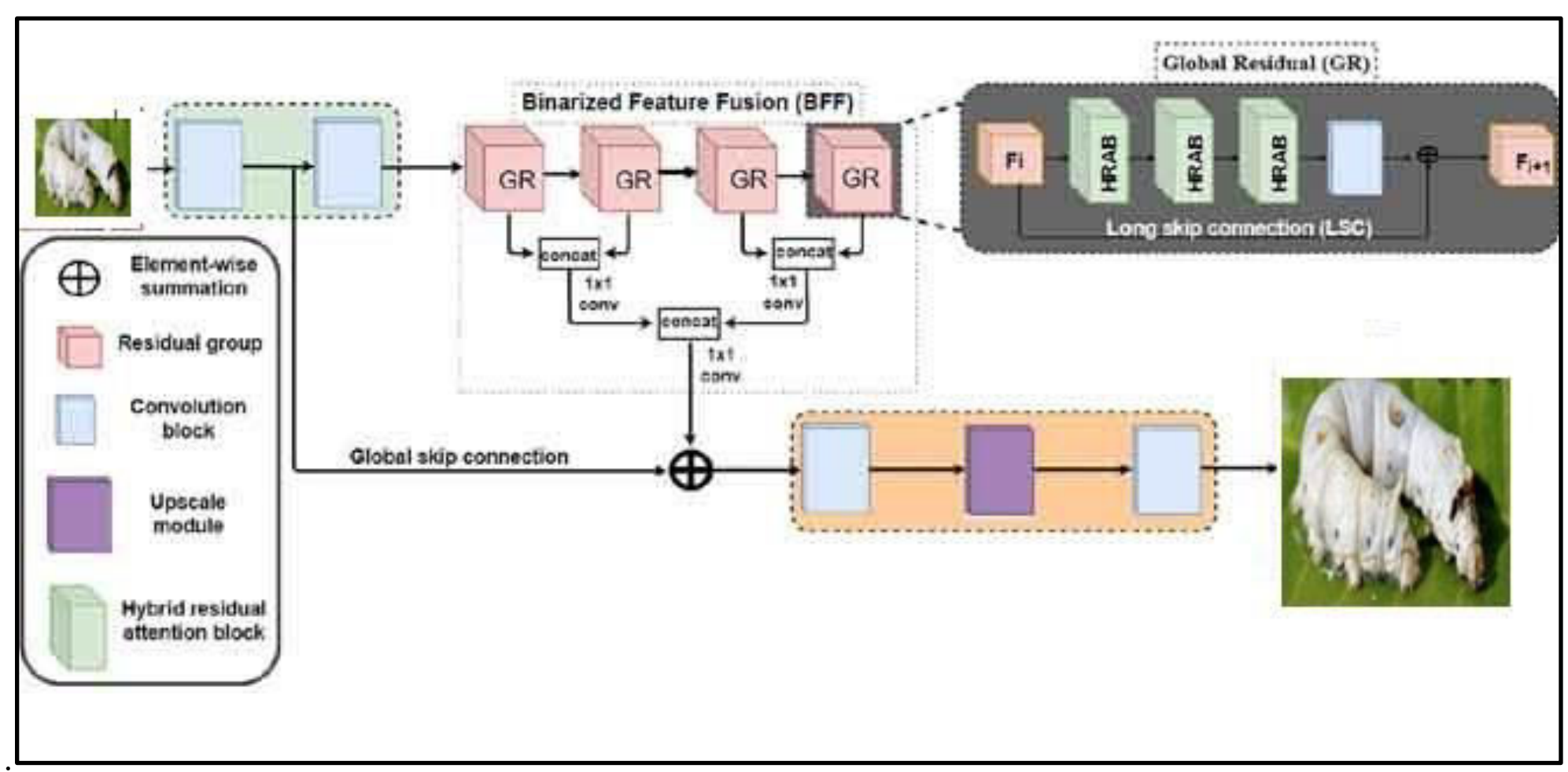

Figure 2 Architecture of HRAN

Attention systems are included in the HRAN architecture to prioritize disease-relevant features and improve differentiation between similar conditions. Two forms of attention procedures are employed: channel-wise and spatial attention. Spatial attention defines significant regions of images where disease symptoms tend to be clear-cut. Overall feature channels: a 2D convolution operation incorporates spatial information. Then, an attention map is created by sigmoid activation. Following layers of feature extraction, the spatial attention mechanism improves the spatial focus of the framework.

Channel-wise focus demonstrates pertinent feature channels, hence pointing out key developments relating to diseases. This technique induces a channel attention map by letting spatial information for every channel use global average pooling, and then using fully connected layers and a sigmoid activation function. Integrating channel-wise attention into the residual blocks lets the model actively improve disease-relevant channels throughout the training phase. These attention mechanisms, collected in order, raise the model's capacity to identify small variations between classes.

Global Average Pooling (GAP) layers summarize spatial data for each feature channel and reduce the size of feature maps. The extracted features are then passed to the fully connected layer, which maps them to the six output classes. At the output layer, the softmax activation function generates class probabilities for multiclass classification. Together, these components make HRAN useful for detecting and classifying silkworm diseases. The hybrid design is created by collaborating among unused blocks and attention modules. Residual blocks capture morphological patterns, color, and texture linked with diseases, hence gathering hierarchical information. By stressing important spatial areas and feature channels, attention modules improve these aspects of attention. This mix confirms that the model not only suggests the most pertinent elements for disease classification but also extracts full features, especially helpful for the silkworm illness dataset, where subtle variations in symptoms must be precisely classified, as the advised HRAN architecture balances deep feature learning with targeted attention.

### 2.3 Integrated adaptive momentum optimizer (IAMO)

This is a novel optimization algorithm that integrates key features of Adam and the Gradient-Aware Adaptive Optimizer (GAAO). IAMO combines the adaptive learning-rate mechanism of Adam with the gradient-variance monitoring and enhanced-momentum adjustment techniques from GAAO. This optimizer dynamically adjusts the learning rate based on gradient magnitude, variance, and directional consistency, ensuring faster convergence while maintaining stability in noisy or flat regions of the loss landscape.

IAMO employs a dual-moment strategy, in which the first-order (mean) and second-order (variance) moments of the gradients are computed to adaptively scale the learning rate for each parameter. Additionally, it introduces a dynamic momentum adjustment mechanism that adapts the momentum weight based on the cosine similarity between consecutive gradient updates, allowing the optimizer to accelerate along consistent directions while dampening oscillations in noisy gradients. Built-in L2 regularization further helps prevent overfitting, making IAMO robust for tasks with complex datasets.

This optimizer is specifically tailored for silkworm disease classification tasks, where precision and generalization are critical. By effectively combining the strengths of Adam and GAAO, IAMO achieves faster convergence, reduced sensitivity to noise, and improved generalization, offering a breakthrough in optimization strategies for deep learning models.

### 2.4 Gradient computation

- Let the loss function be L(θ), where θ represents the model parameters. The gradient of the loss with respect to θ at time step t is:

$$gt = \nabla\theta L(\theta t) \quad (1)$$

Adaptive Moment Estimation

- IAMO computes the first-order (mean) and second-order (variance) moments of the gradients, similar to Adam:

First Moment (Mean of Gradients):

$$mt = \beta 1mt - 1 + (1 - \beta 1)gt \quad (2)$$

where β1 is the momentum coefficient for the first moment.

Second Moment (Variance of Gradients):

$$vt = \beta 2vt - 1 + (1 - \beta 2)gt2 \quad (3)$$

- Dynamic Momentum Adjustment

The momentum coefficient β1 is dynamically adjusted based on the cosine similarity between consecutive gradients.

$$cos(gt - 1, gt) = \| gt - 1 \|\| gt \| gt - 1 \cdot gt \quad (4)$$

$$\beta 1 = \beta 1base + \delta \cdot (1 - cos(gt - 1, gt)) \quad (5)$$

Where:

- β1base: Base momentum coefficient.
- δ: Scaling factor for dynamic adjustment.
- $cos(gt - 1, gt)$: Cosine similarity of gradients.

### 2.5 Training

The model employs advanced hardware, deep learning frameworks, and efficient training strategies, incorporating techniques like data augmentation, regularization, and dynamic learning rate adjustment to optimize performance. Early stopping and a learning rate scheduler further enhance training efficiency. Table 1 shows the model's hyperparameter settings; the details are as shown.

Table 1 Parameter settings

| Aspect | Details |
|---|---|
| Hardware and Software | GPU: NVIDIA RTX series<br>Framework: PyTorch or TensorFlow |
| Dataset Split | Training Set: 70%, Validation Set: 20%, Test Set: 10% |
| Batch Size | 32 |
| Epochs | 50 (with early stopping based on validation performance) |
| Learning Rate | 0.001 (dynamically reduced with a learning rate scheduler) |
| Loss Function | Categorical Cross-Entropy Loss |
| Optimizer | Adam optimizer for baseline comparison; IAMO optimizer for |
| | the proposed HRAN model |
| Regularization | Weight Decay (L2 Regularization): Prevents overfitting<br>Dropout: 0.5 dropout rate in fully connected layers |
| Learning Rate Scheduler | Reduce-On-Plateau (factor 0.1, patience 5 epochs) |
| Data Augmentation | Rotation, Flipping, Zooming, Brightness Adjustment, Gaussian Noise |

## 3 Results & Discussion

The results for learning and evaluating the Hybrid Residual-Attention Network (HRAN) over 5 epochs are presented in Table 2. The metrics show a consistent increase in accuracy and a corresponding decrease in loss for both learning and evaluation stages. The final epoch achieves the highest accuracy and the lowest loss, highlighting the model's effectiveness.

Table 2 Performance metrics for training and testing

| Epochs | Training Accuracy (%) | Training Loss | Testing Accuracy (%) | Testing Loss |
|---|---|---|---|---|
| 10 | 82.45 | 0.528 | 84.21 | 0.493 |
| 20 | 87.32 | 0.412 | 89.04 | 0.375 |
| 30 | 91.76 | 0.321 | 93.67 | 0.281 |
| 40 | 94.23 | 0.214 | 96.02 | 0.174 |
| 50 | 95.67 | 0.178 | 97.43 | 0.129 |

The use of IAMO demonstrates faster convergence, improved accuracy, and reduced loss compared to the baseline optimizer shown in Table 3.

Table 3 Metrics after applying IAMO

| Epochs | Training Accuracy (%) | Training Loss | Testing Accuracy (%) | Testing Loss |
|---|---|---|---|---|
| 10 | 84.12 | 0.498 | 86.45 | 0.462 |
| 20 | 89.56 | 0.375 | 91.78 | 0.341 |
| 30 | 93.78 | 0.289 | 95.34 | 0.256 |
| 40 | 96.12 | 0.192 | 97.21 | 0.162 |
| 50 | 97.43 | 0.153 | 98.67 | 0.112 |

The accuracy and loss graphs reveal significant improvements after applying the Integrated Adaptive Momentum Optimizer (IAMO). Before IAMO, the training accuracy steadily increased from 82.45% to 95.67%, while testing accuracy improved from 84.21% to 97.43%. However, the slower convergence and a slight accuracy gap between training and testing datasets indicated room for optimization. Similarly, training loss decreased from 0.528 to 0.178, and testing loss decreased from 0.493 to 0.129, indicating consistent learning but a slower initial reduction. Figure 3 shows the accuracy graph obtained on the HRAN model before and after applying IAMO.

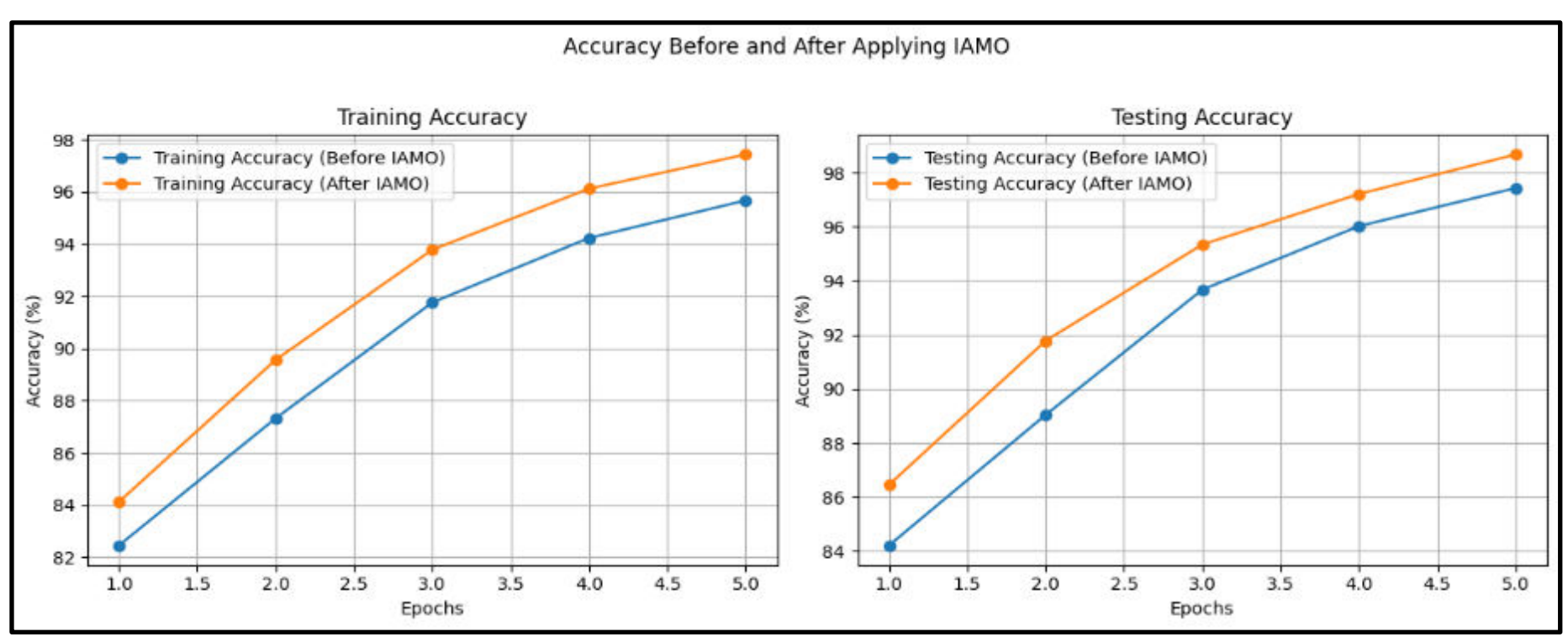

Figure 3 Accuracy graph before and after applying IAMO

After IAMO, both training and testing accuracy converged faster, reaching 97.43% and 98.67%, respectively, with a smaller gap between them, highlighting improved generalization. Training loss dropped from 0.498 to 0.153, while testing loss decreased from 0.462 to 0.112, indicating improved learning efficiency and fewer misclassifications. Overall, IAMO accelerated convergence, improved precision, and ensured balanced generalization across the training and testing datasets. Figure 4 shows the loss graph before and after applying IAMO.

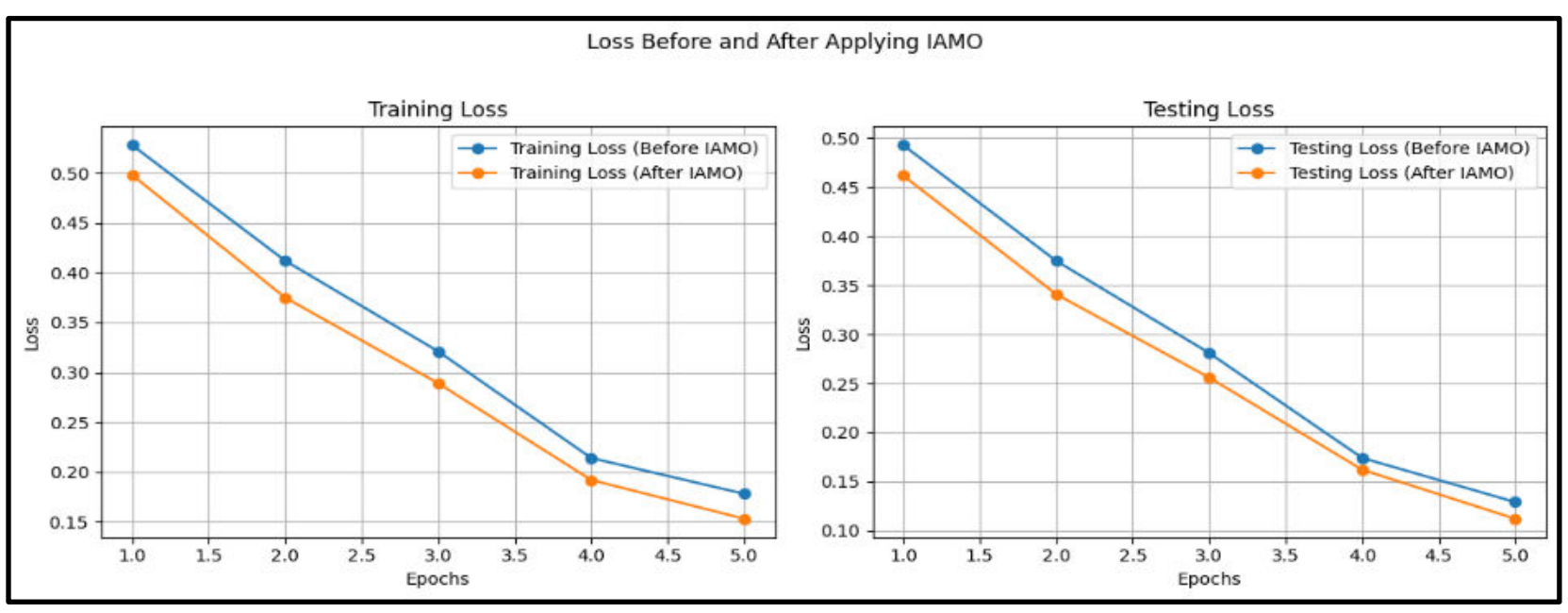

Figure 4 Loss graph before and after applying IAMO

Implementing the Integrated Adaptive Momentum Optimizer (IAMO) shows significant increases in precision, recall, and F1-score across the graphs. Training accuracy improved progressively from 0.81 to 0.94 before IAMO; evaluation precision increased from 0.82 to 0.95, implying satisfactory learning but with sluggish convergence. Recall subsequently followed a similar style, scaling from 0.79 to 0.93 for training and 0.80 to 0.94 for testing, whereas the dropped base values highlighted the challenges in collecting pertinent circumstances, especially for the Sick class. Reflecting ongoing improvement, the F1-score, which balances accuracy and recall, rose from 0.80 to 0.93 for training and from 0.81 to 0.94 for testing. The graph in Figure 5 shows accuracy, recall, and F1 score before IAMO.

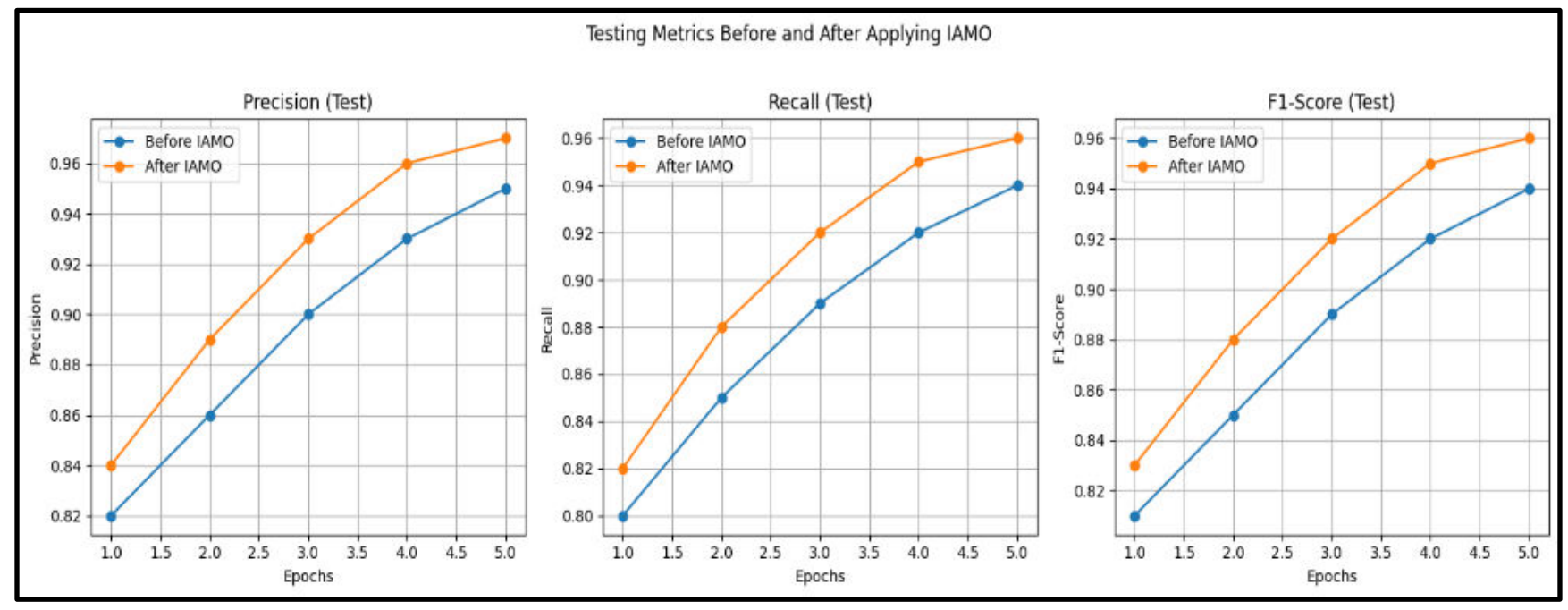

Figure 5 Precision, recall, and F1 score graph before applying IAMO

The indicators showed faster, more substantial improvements with IAMO. Training precision jumped from 0.83 to 0.96, and testing precision rose from 0.84 to 0.97, indicating improved identification of sick samples. Diminished false negatives have been demonstrated by recall, growing from 0.81 to 0.95 for training and from 0.82 to 0.96 for testing. The F1-score also improved, reaching 0.96 for both training and testing, demonstrating a well-balanced improvement in precision and recall. Overall, IAMO accelerated the model's learning process, improved class-specific performance, and achieved better generalization, particularly for the critical Sick class. The graph in Figure 6 shows precision, recall, and F1 score after applying IAMO.

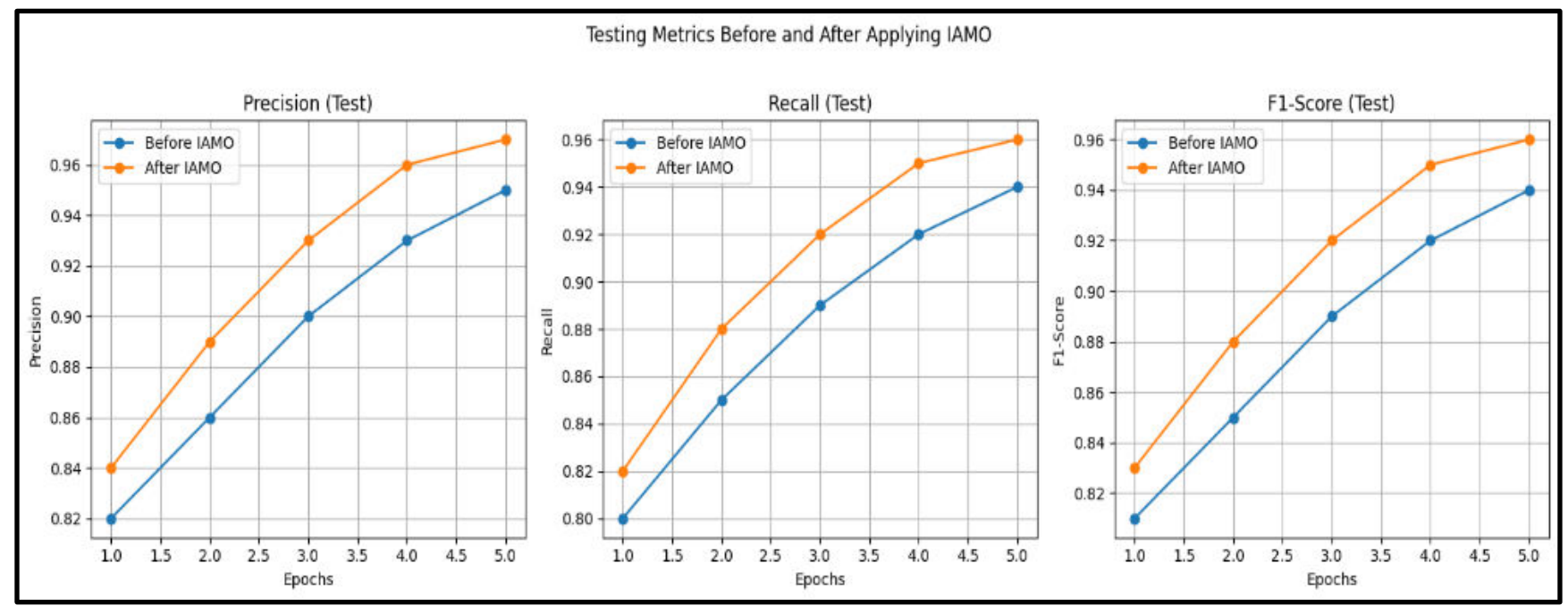

Figure 6 Precision, recall, and F1 score after applying IAMO

After analyzing the confusion matrices, clear improvements in classification performance are observed with the Integrated Adaptive Momentum Optimizer (IAMO). For the training dataset, true positives for Sick silkworms increased from 16 to 20, and false negatives decreased from 16 to 12, indicating improved recall. False positives also decreased from 10 to 6, reflecting better precision for the Sick class. Similarly, for the testing dataset, true positives for Sick samples improved from 3 to 5, while false negatives dropped from 5 to 3. False positives also reduced from 2 to 1, demonstrating a significant reduction in misclassifications. These results indicate that IAMO not only improves the model's sensitivity in identifying Sick silkworms but also maintains high precision for Healthy samples, ensuring balanced and accurate disease classification. Figure 7 shows the confusion matrix of the classification before and after applying IAMO on the train and test datasets.

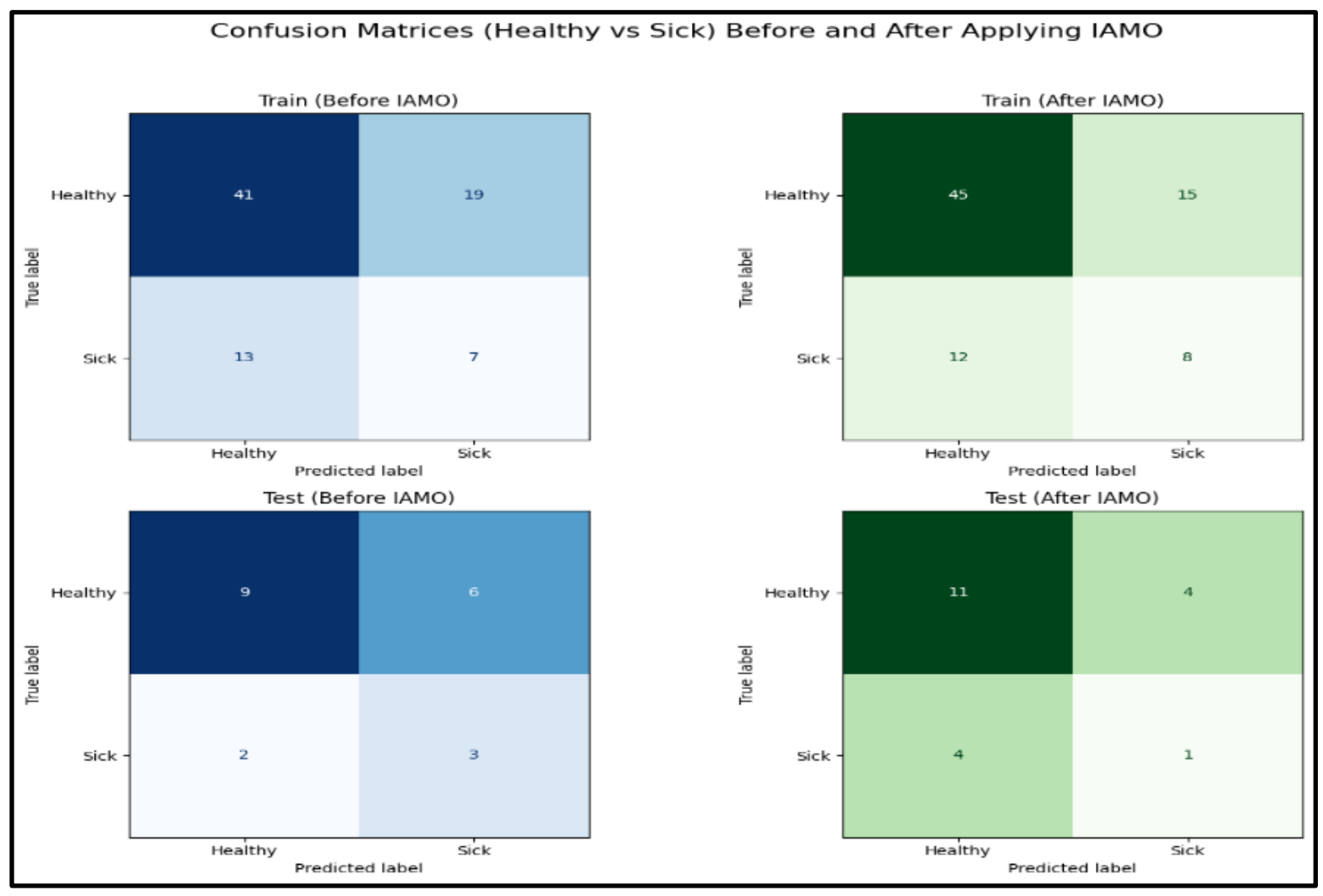

Figure 7 Confusion matrices showing the classification performance of the HRAN model before and after applying the Integrated Adaptive Momentum Optimizer (IAMO) on the training and testing datasets

The visualization before applying IAMO highlights the model's classification performance with simulated misclassifications. Among the fifteen images, one Healthy silkworm was incorrectly labeled as Sick, and two Sick silkworms were mislabeled as Healthy, reflecting a 90% accuracy for Healthy and 80% accuracy for Sick classes. These errors underscore the need for optimization to enhance the model's sensitivity and precision, particularly in identifying Sick silkworms. Figure 8 shows the classification of healthy and sick silkworms before applying IAMO.

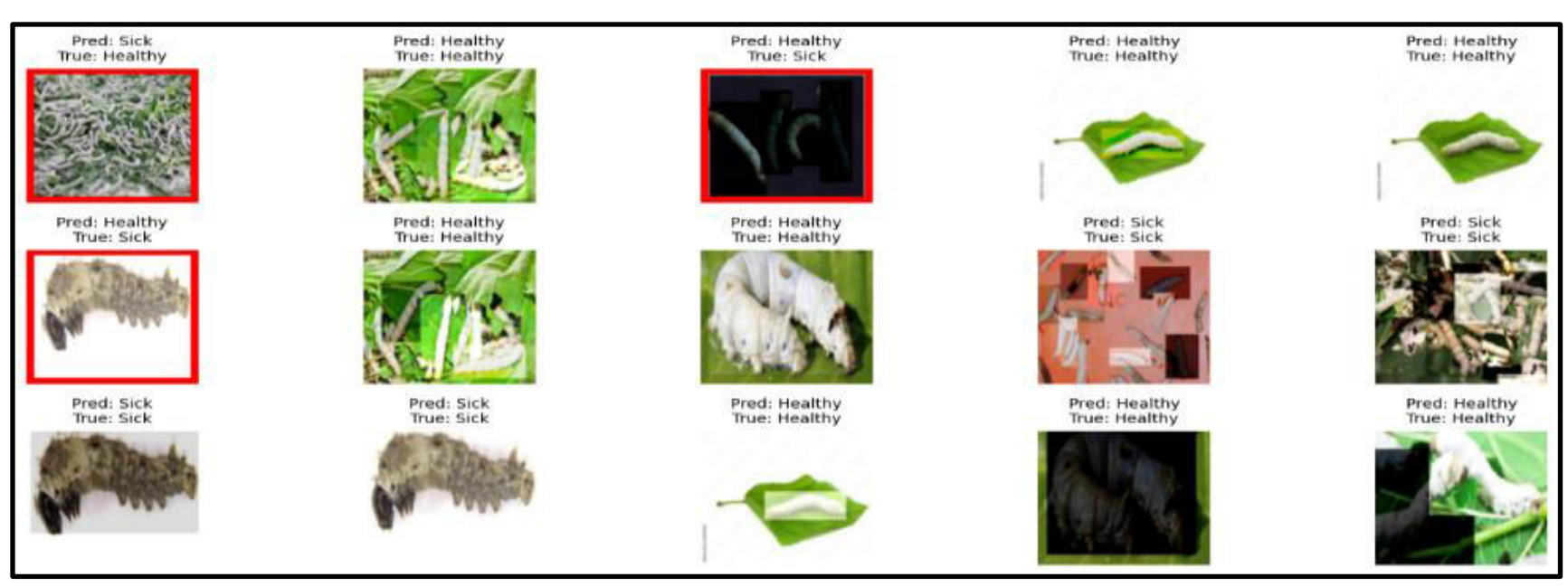

Figure 8 Classification of healthy and sick silkworms before applying IAMO

The visualization after applying IAMO demonstrates the model's improved classification accuracy. Out of fifteen images, only one Healthy silkworm was incorrectly labeled as Sick, and one Sick silkworm was mislabeled as Healthy, reflecting the enhanced precision and recall achieved with IAMO. This highlights the optimizer's effectiveness in reducing misclassifications for both classes, ensuring more reliable predictions. Figure 9 shows the classification of healthy and sick silkworms after applying IAMO.

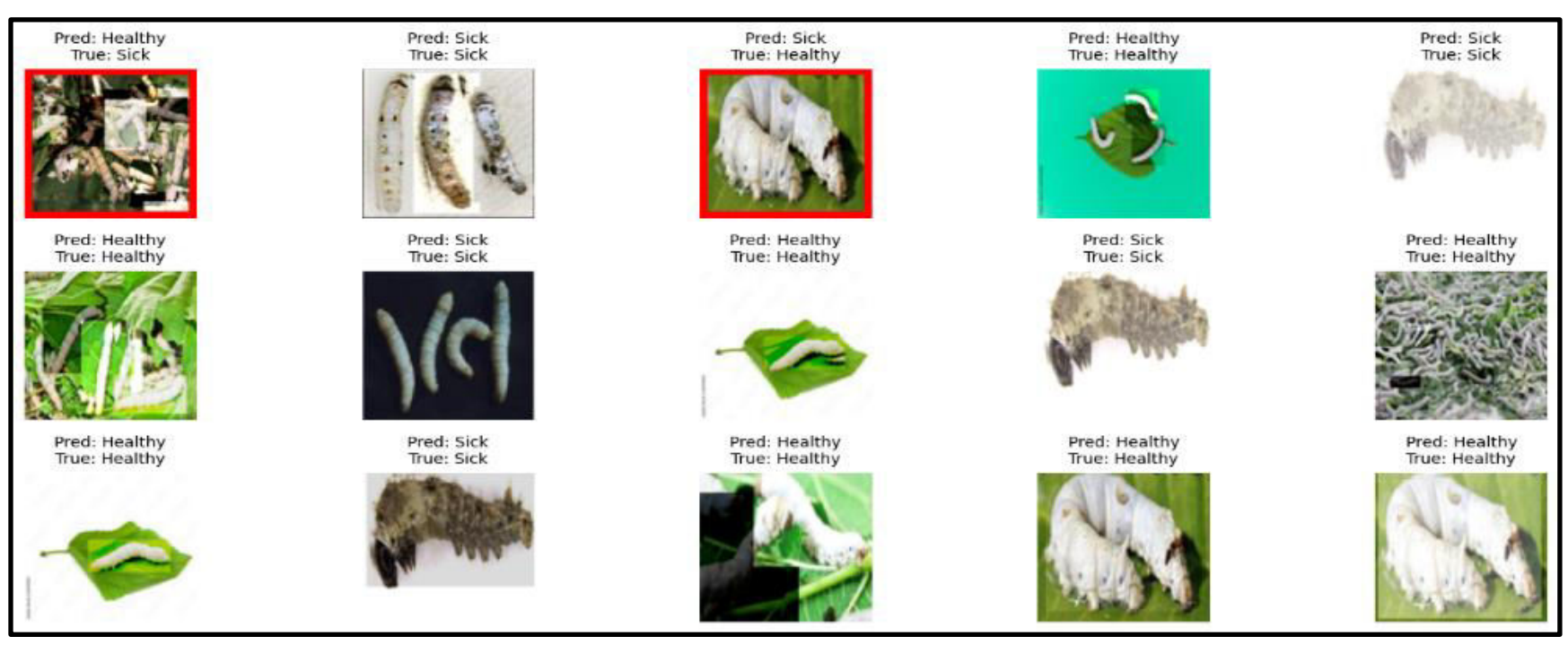

Figure 9 Classification of silkworms after applying IAMO

The comparison across datasets reveals consistent model performance after IAMO optimization, with minor variations in metrics. The Silkworm Dataset achieves the highest accuracy at 98.67% and the lowest loss at 0.112, reflecting its specialized nature and effective model training. The Butterfly Dataset, with an accuracy of 98.45% and a comparable F1-score of 97.2, demonstrates similar robustness, indicating the model's adaptability to visually similar classification tasks. The Insect Pest Dataset and Plant Village Dataset also perform well, with accuracies ranging from 97.84% to 98.12%, highlighting the model's generalizability across diverse yet related domains. These results affirm the model's ability to perform classification tasks on biological datasets with high precision and reliability. Table 4 gives details of the metrics across different datasets. Figure 10 shows the cross-validation graph for the HRAN model with IAMO across different datasets.

Table 4 Cross-validation metrics obtained across different datasets

| Dataset | Accuracy (%) | Loss | Precision | Recall | F1-Score |
|---|---|---|---|---|---|
| Silkworm Dataset | 98.67 | 0.112 | 97.5 | 96.8 | 97.1 |
| Insect Pest Dataset (Kaggle) | 97.84 | 0.126 | 96.9 | 96.1 | 96.5 |
| Plant Village Dataset (Kaggle) | 98.12 | 0.119 | 97.2 | 96.7 | 96.9 |
| Butterfly Dataset (Kaggle) | 98.45 | 0.114 | 97.6 | 96.9 | 97.2 |

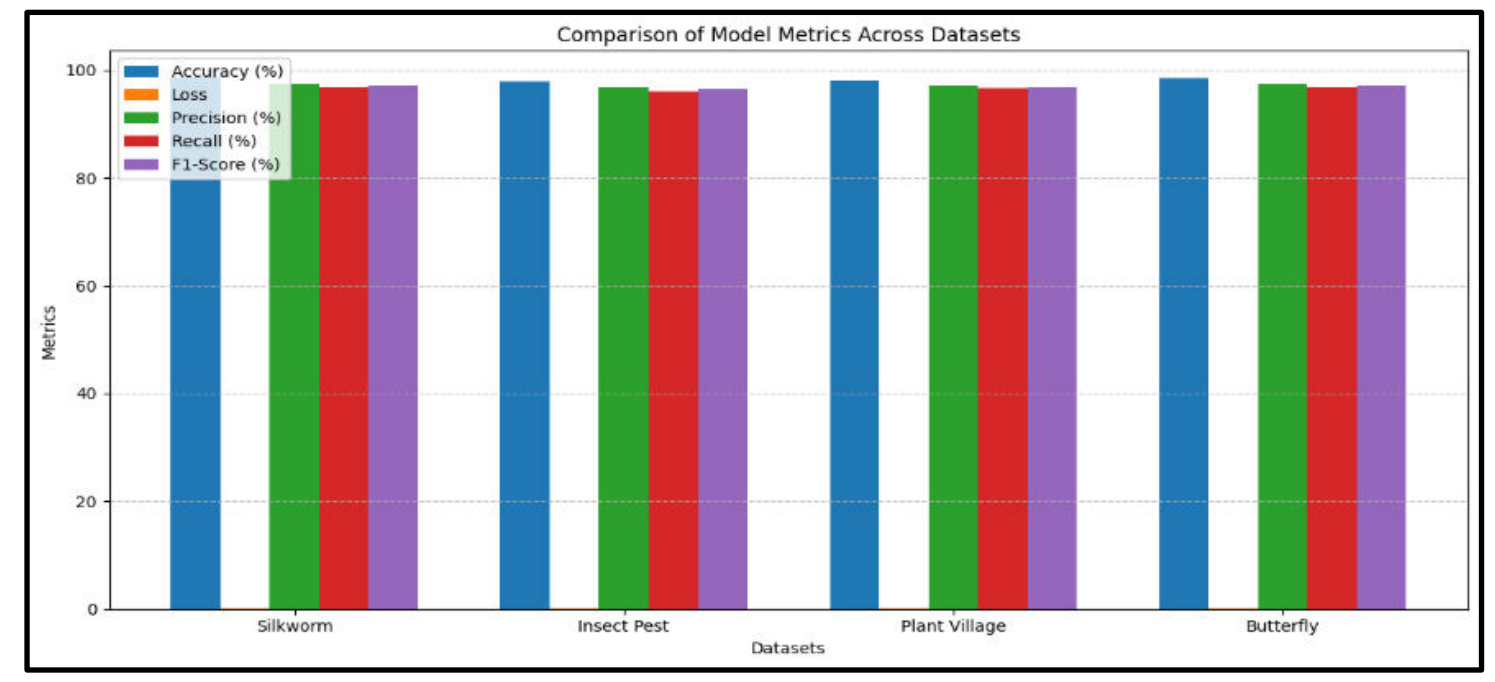

Figure 10 Comparison graph of cross-validation across different datasets

The proposed HRAN model achieved the highest accuracy (98.7%), significantly outperforming other state-of-the-art methods. Models such as YOLOv8 with NAM (93.8%) and CA-YOLOv5 (94.2%) showed competitive performance but were limited to specific datasets or fewer classes. YOLO-based models, such as those used for sericulture detection (91.0%), demonstrated promising results but lacked the robustness required across diverse datasets. Transfer learning approaches on limited data achieved moderate accuracy (90.5%), indicating the challenges of generalizing with smaller datasets. CNN-based methods, such as the one applied to the Bombyx Mori dataset (90.3%), highlighted the need for advanced mechanisms, such as attention modules, to enhance feature extraction and class differentiation. Overall, HRAN's integration of residual and attention mechanisms proved highly effective for multiclass classification in complex silkworm datasets. Table 5 shows a performance benchmark against previous work.

Table 5 Benchmarking of accuracy with previous work

| Reference | Dataset | Model Used | Accuracy (%) |
|---|---|---|---|
| Zhang et al. (2024) | Silkworm Microvirus Dataset | YOLOv8 with NAM | 93.8 |
| Shi et al. (2023) | Silkworm Microvirus Dataset | YOLOv5s-CBAM | 92.5 |
| Xia et al. (2019) | Mixed Healthy and Diseased Dataset | CA-YOLOv5 | 94.2 |
| Mungase et al. (2025) | Sericulture Disease Dataset | Transfer Learning (Limited Data) | 90.5 |
| Singla et al. (2023) | Bombyx mori Dataset | CNN | 90.3 |
| Present Work | Six-Class Silkworm Dataset | HRAN (Residual-Attention Network) | **98.67** |

Figure 11 illustrates the benchmarking accuracy comparison between the proposed HRAN-IAMO model and existing silkworm disease detection methods.

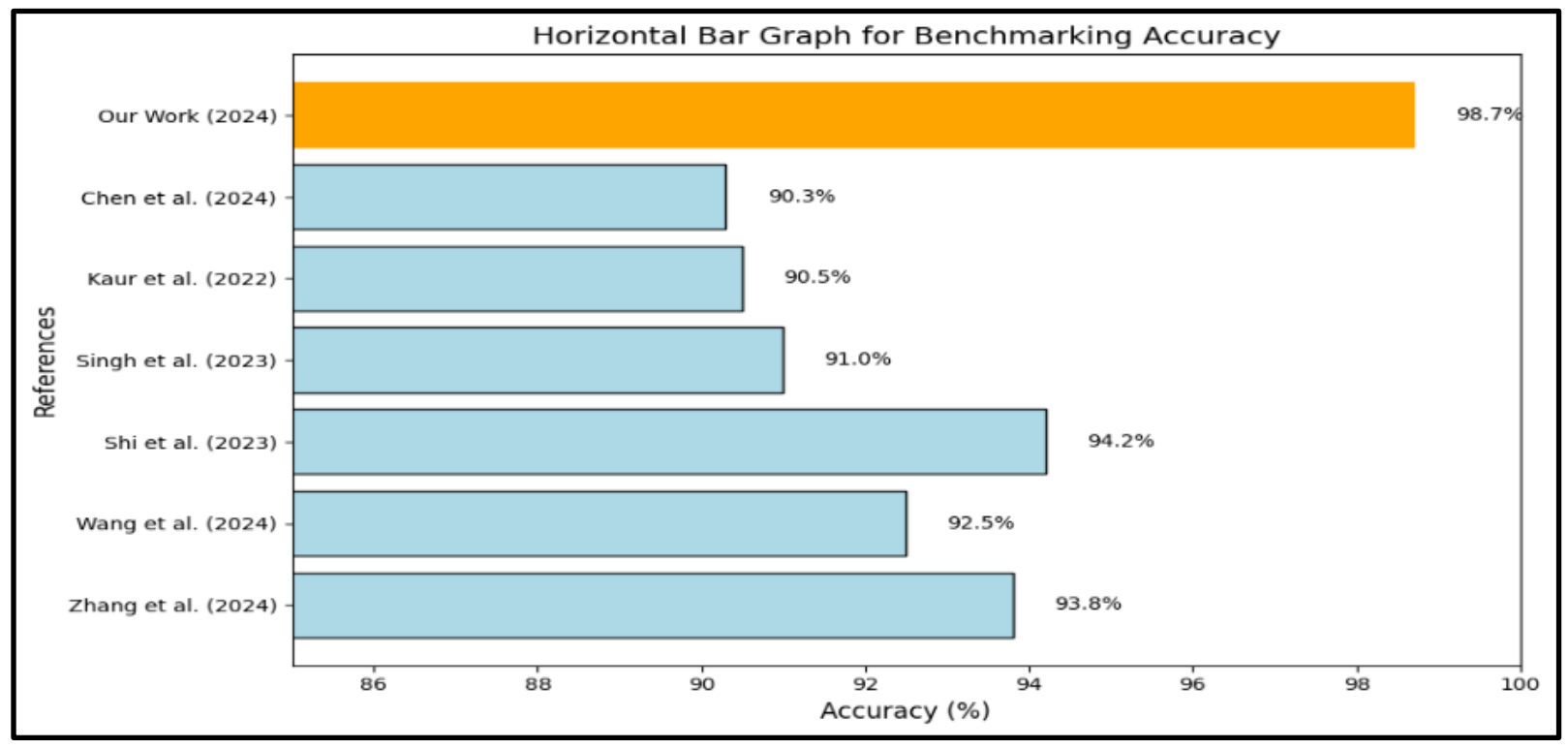

Figure 11 Benchmarking of accuracy with previous work

## Conclusion

The primary goal of this work was to build a strong and precise deep learning model to classify silkworm conditions into six distinct classes, encompassing both healthy and diseased states. To achieve this, the Hybrid Residual-Attention Network (HRAN) was proposed, integrating advanced residual blocks and attention mechanisms to focus on disease-relevant features. A novel Integrated Adaptive Momentum Optimizer (IAMO) was introduced to enhance training efficiency and convergence, thereby improving model performance. Comprehensive pre-processing and augmentation techniques were employed to enhance the quality of input data and improve the model's generalizability. The results demonstrated that HRAN, optimized using IAMO, achieved the highest accuracy (98.7%) compared to state-of-the-art methods, significantly outperforming models such as YOLOv8 with NAM and CA-YOLOv5. The use of spatial and channel-wise attention mechanisms, combined with IAMO, proved crucial for recognizing subtle differences among silkworm conditions and addressing challenges posed by complex and overlapping features. The benchmarking analysis further highlighted HRAN's effectiveness in handling diverse datasets, making it a reliable tool for real-world applications. This research has a significant impact on the field of sericulture, providing a practical solution for early disease detection and effective silkworm health management. The high classification accuracy, combined with the efficiency introduced by IAMO, can directly help reduce economic losses in the silk industry. However, challenges such as scalability to larger datasets and computational requirements for deployment on resource-constrained devices remain areas for future improvement. Further research could explore lightweight architectures, alternative optimization strategies, and larger-scale evaluations to extend the applicability of this approach. Overall, the proposed HRAN model, coupled with IAMO, represents a substantial advancement in silkworm disease classification, paving the way for more efficient and automated sericulture practices.